\documentclass[journal]{IEEEtran}
\usepackage{graphicx}
\usepackage{paralist}
\usepackage[inkscapelatex=false]{svg}
\usepackage[shortlabels]{enumitem}
\usepackage{hyperref}
\usepackage{multirow}
\usepackage{dblfloatfix} 
\usepackage{amsmath}
\usepackage{mathptmx}
\usepackage{tabularx}
\usepackage{algorithm}
\usepackage[noend]{algpseudocode}
\usepackage{lineno}
\usepackage{comment}
\usepackage{amssymb}
\usepackage{microtype}
\usepackage{bm}
\usepackage{colortbl}
\usepackage{cite}
\usepackage{svg}

\begin{document}

\title{Dynamic Human-Robot Role Allocation through Entropy-Based Monitoring and LLM Reasoning}

\author{Elena~Merlo, Marta~Lagomarsino, Hamidreza~Raei, and Arash~Ajoudani
\thanks{This work was supported in part by the Italian Ministry of University and Research (MUR) under the Fondo Italiano per la Scienza (FIS), call FIS 3, project EPIC with code FIS-2024-02654, and by the HE project TORNADO (GA 101189557). All the authors are with Human-Robot Interfaces and Interaction, Istituto Italiano di Tecnologia, Genoa, Italy (e-mail: elena.merlo@iit.it, marta.lagomarsino@iit.it, hamidreza.raei@iit.it, arash.ajoudani@iit.it). Hamidreza Raei is also with Department of Electronics, Information and Bioengineering, Politecnico di Milano, Milan, Italy.}
\vspace{-1.2cm}}


\maketitle

\begin{abstract}
Role allocation strategies in human-robot collaboration typically assign tasks to both the human and the robot, limiting the user's ability to take initiative. Furthermore, allocation decisions may conflict with the user's intentions, and the rationale behind such allocations may remain unclear, potentially reducing both acceptance and collaboration fluency. 
The novelty of the proposed framework is that the human is left free to contribute to the task, while the robot is dynamically allocated activities through a cost-driven Behavior Tree architecture according to user-expressed preferences and behavior. Explicit preferences are extracted from natural language instructions using a large language model, whereas human activity is monitored online through a Shannon entropy-based analysis of hand-object interactions. Furthermore, the framework can recognize anomalies in human execution and autonomously trigger corrective robot actions. 
The approach was validated through a pilot study and a multi-subject experiment involving 14 participants. Results demonstrate the potential of such an allocation strategy for enabling intuitive and engaging collaboration without increasing the perceived workload.
\vspace{-0.3cm}
\end{abstract}

\begin{IEEEkeywords}
Human-robot collaboration, dynamic role allocation, behavior trees, information theory, large language models.
\vspace{-0.5cm}
\end{IEEEkeywords}

\IEEEpeerreviewmaketitle

\vspace{-0.2cm}

\section{Introduction}
\IEEEPARstart{R}{ole} allocation in human-robot collaboration (HRC) originally relied on fixed, predefined strategies derived from the inherent capabilities of humans and robots. In line with Fitts' List of ``Men Are Better At/Machines Are Better At" \cite{fitts1951human}, humans were typically assigned tasks demanding reasoning, adaptability, or creativity, whereas robots were responsible for operations that were repetitive, precise, or hazardous. 
As technology advances and increasingly powerful AI models emerge, the traditional separation between human and robot capacities is becoming less distinct, calling for adaptive and context-sensitive role assignment strategies \cite{lagomarsino2025intuitive}.
By monitoring online the state of the human, the robot, and the surrounding environment, roles can be dynamically adjusted during task execution, optimizing the team performance and fostering fluid collaboration \cite{petzoldt2023review, lorenzini2023ergonomic}.

In industry and healthcare, repetitive tasks and heavy lifting can lead to musculoskeletal injuries and reduced productivity. Embedding physical condition and ergonomic assessments within role allocation strategies facilitates optimized load sharing, promoting both human well-being and system efficiency.
Classical approaches rely on explicit ergonomic thresholds or rule-based criteria. For example, Rapid Entire Body Assessment (REBA)-based systems \cite{makrini2019task} reassigned tasks to the robot when human posture becomes unsafe, while other methods combined initial offline allocation with online fatigue estimation from motion tracking \cite{calzavara2024achieving}.
To manage more complex constraints, several works use optimization-based role allocation. Mixed-Integer Programming (MIP) formulations \cite{maderna2020online, pupa2021human} balanced makespan, precedence, and ergonomic load from estimated worker's posture, updating task assignments online as human performance evolves. In \cite{merlo2023ergonomic}, authors proposed an AO$^\star$ search over an AND/OR Graph (AOG) of feasible assembly sequences and agent assignments, with nodes as subassemblies and agent-specific edges as actions. Task costs reflect agent suitability: robots have fixed costs and penalized infeasible tasks, while human costs depend on KWear, derived from Rapid Upper Limb Assessment (RULA) scores and posture duration. When KWear exceeds safe limits, tasks shift to the robot until recovery.
Alternatively, data-driven approaches can estimate fatigue through learned biomechanical models \cite{messeri2022dynamic}, assigning demanding tasks to the robot. When tasks cannot be decomposed, allocation shift to the control level, with robots adapting assistance policies based on human fatigue measured via EMG \cite{peternel2018robot} or switching control modes dynamically \cite{vianello2024effects}: a strategy that is effective but often task-specific and difficult to generalize.
However, although these strategies reduce musculoskeletal strain, when tasks have similar ergonomic risk scores, the resulting role assignment may become non-intuitive. While users readily accept not being assigned to clearly high-risk tasks like heavy lifting, being asked to perform a task with comparable risk but against their intentions can also feel frustrating.

\begin{figure*}
    \centering
    \includegraphics[width=0.9\linewidth]{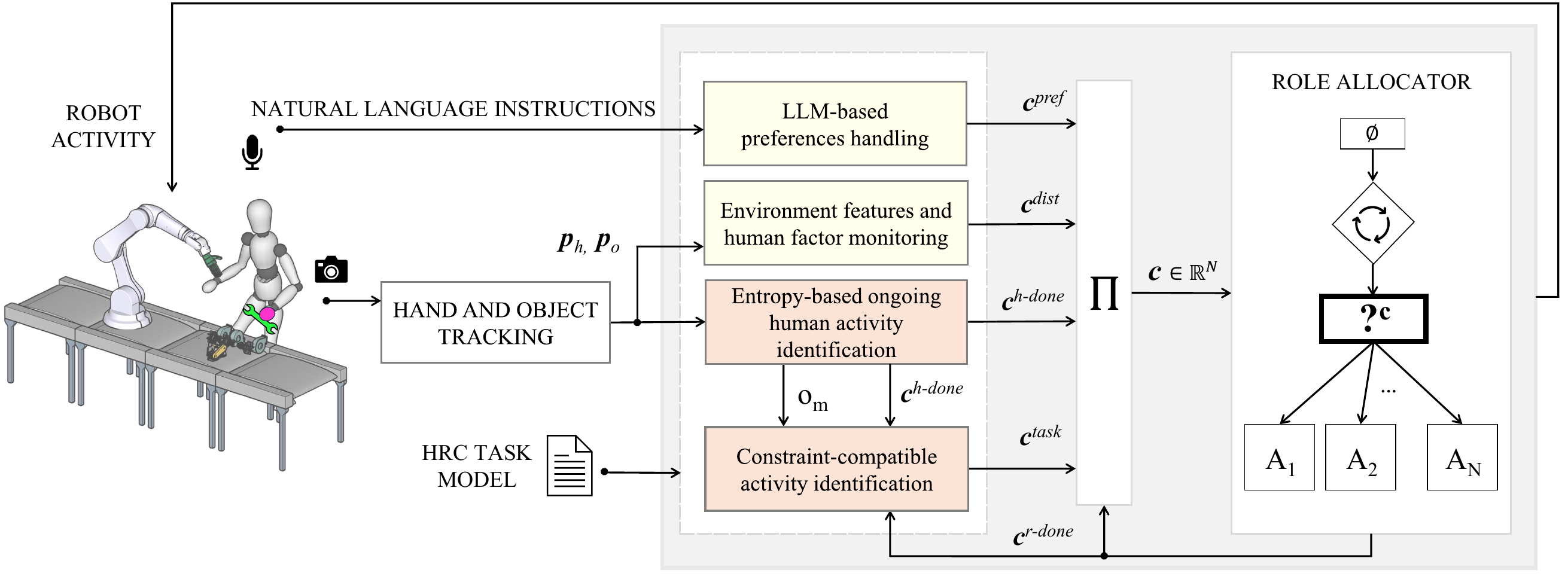}
    \vspace{-0.4cm}
    \caption{Overview of the proposed framework: LLM-interpreted user preferences, human factor monitoring, human activity identification, and task constraints inspection are integrated to form a cost vector $\boldsymbol{c}$ used by the designed \texttt{Cost-Based Fallback} node (symbol $\text{?}^\textbf{c}$) in a Behavior Tree, enabling the user to act freely while the robot adapts accordingly selecting the activity $A_i$ to execute. The costs deriving from the orange modules encode which are the available activities, while those from the yellow modules indicate which ones to prioritize.}
    \label{fig:role_alloc_framework}
    \vspace{-0.5cm}
\end{figure*}

Other approaches exploit cognitive load measures or user preferences to improve both performance and user satisfaction. Early approaches relied on rule-based strategies: the robot adjusts its autonomy level or reassigns subtasks when cognitive load, stress, or trust estimates cross predefined thresholds \cite{giele2015dynamic, wang2023robot}.
A second family of methods formulated allocation as an optimization problem, selecting roles that balanced productivity with human well-being, keeping mental or physiological stress within safe limits \cite{lagomarsino2025promind, rahman2018mutual}.
More recent work has shifted toward learning-based personalization. Robots learn human capabilities or trustworthiness from task outcomes and allocate roles accordingly \cite{ali2022heterogeneous}, or model collaboration as a two-player game in which both partners negotiate roles that satisfy productivity and stress objectives, approaching equilibrium through experience \cite{messeri2021human}.
In this context, estimates of trust or stress are not intentional inputs provided by the user, so the resulting allocations may confuse an operator who still wishes to retain some freedom in choosing what to do. More recently, Large Language Models (LLM)s have been used to process explicit user requests, which represents a step toward more intuitive collaboration thanks to the naturalness of verbal communication.
For instance, \cite{izquierdo2024plancollabnl} allowed users to express states or preferences directly (e.g., \textit{``I have a back pain"}), which an LLM translated into allocation costs, reducing the likelihood of assigning to the user tasks that strain their back.

While the above-mentioned approaches typically prioritized a single optimization objective, the following combine multiple factors simultaneously.
\cite{lamon2019capability} introduced a capability-based framework scoring agent-task pairs by task complexity, dexterity, and physical effort.
\cite{lippi2023task} formulated allocation as a Mixed-Integer Linear Programming (MILP) problem that accounts for makespan, workload, quality, preferences, and spatiotemporal constraints, with dynamic re-allocation during execution, but its computational burden limits online use.
To address this, \cite{fusaro2021integrated} leveraged Behavior Trees (BT)s to decompose jobs into smaller subproblems, solving a MILP only for currently unlocked tasks. Allocation considered capabilities, ergonomics, and availability.
However, having many factors influencing the allocation decision, it becomes even harder for the user to understand why certain tasks are assigned to them.
In \cite{lamon2023unified}, an augmented reality interface was used for negotiation: the user could explicitly accept or reject proposed allocations, updating preference-based costs that influenced future algorithm decisions. In this way, the user became part of the decision-making process, recognizing that their intentions were taken into account.
However, continuously requiring the user to confirm allocations can negatively affect the fluency of the interaction, interrupting task execution and eventually becoming tedious and unnatural \cite{candon2026learning}. Moreover, the human freedom of action remains limited, as the user can only accept or reject the proposed allocation without actively taking initiative within the collaboration. 
Allowing users to choose actions on the fly, rather than following imposed allocations, remains relatively underexplored despite its relevance for tasks admitting multiple valid execution sequences, such as lightweight assembly. To include human decision-making authority in a role allocation framework, \cite{darvish2020hierarchical} used wearable devices to recognize gestures, informing an AOG-based task manager of the human's ongoing activity and triggering compatible robot actions. However, their approach required task-specific gesture training and did not assess activity completion and correctness.

This work proposes a dynamic role allocation framework that aims to promote both \textit{intuitiveness} and \textit{human agency} in HRC tasks. 
To foster \textit{intuitive} interaction, users can communicate with the robot through natural language, as they would with a human teammate. By leveraging the commonsense reasoning capabilities of an LLM, the robot adapts its behavior to the provided instructions, making its actions easier to anticipate and understand.
To support \textit{human agency}, our framework monitors hand-object interactions through an RGB camera and evaluates their information flow using Shannon entropy measures \cite{shannon1948mathematical}, representing the evolution of human activity through dynamic scene graphs \cite{merlo2025exploiting} and allowing its identification without any task-specific training phase. Moreover, the structural properties of the generated scene graphs allow the system to assess whether an activity has been completed correctly, providing a basis for detecting execution anomalies and triggering appropriate robot interventions.
The main methodological contributions are the design of:
\begin{inparaenum}[(i)]
\item a cost-driven role allocation framework in which heterogeneous factors describing the \textit{collaboration status}, including spoken instructions, are encoded as cost vectors and integrated within a dedicated BT-based decision-making structure;
\item a Shannon entropy-based algorithm for identifying the activities the user spontaneously decides to perform and evaluating the correctness of the execution, starting from the online version of \cite{merlo2025exploiting}.
\end{inparaenum}
An overview of the proposed framework is shown in Fig.~\ref{fig:role_alloc_framework}.

The remainder of the manuscript is organized as follows. Section II details the design of the BT-based role allocator and the involved cost computation. Section III presents the experimental campaign conducted to validate the framework and evaluate multi-subject experience. Section IV reports the experimental results. Section V concludes the manuscript discussing results and framing limitations and future directions.

\vspace{-0.3cm}
\section{Method}
The proposed framework is presented in three steps. First, the BT-based role allocator is introduced as the central decision-making layer that integrates the cost information describing the current \textit{collaboration status} and selects the activity to be executed by the robot (Section \ref{ssec:role_allocator}). Then, the modules computing the different cost terms are detailed (Section \ref{ssec:costs_computation}). Finally, the entropy-based method designed for online human activity monitoring and recognition is described (Section \ref{ssec:entropy}).
Section \ref{ssec:example_alloc} provides an explanatory example of how the overall allocation process works during HRC task execution.

\vspace{-0.3cm}
\subsection{Role Allocator}
\label{ssec:role_allocator}
BTs are directed rooted trees widely adopted in robotics for task execution and online decision-making due to their modularity, reactivity, and flexibility. Execution starts from the root node, which periodically propagates a signal, called \textit{tick}, to its children. Once ticked, each node returns one of three possible statuses to its parent: \texttt{RUNNING}, \texttt{SUCCESS}, or \texttt{FAILURE}. 
BT nodes can be divided into execution nodes and control nodes. Execution nodes include \textit{action} nodes, which trigger robot primitives, and \textit{condition} nodes, which evaluate logical conditions online. Control nodes instead regulate the execution flow among their children. Among the most common control nodes, a \texttt{Sequence} node ticks its children in order and returns \texttt{SUCCESS} only if all of them succeed, whereas a \texttt{Fallback} node acts as an OR operator, sequentially ticking its children until one returns \texttt{SUCCESS}. Another class of control nodes is represented by \textit{decorator} nodes, which modify the execution policy or returned status of a single child node.

To keep the role allocation process continuously active throughout the collaboration, the proposed BT integrates a \texttt{KeepRunningUntilFailure} decorator connected to the root node. This decorator repeatedly ticks its child subtree until a \texttt{FAILURE} status is returned, allowing the robot to continuously re-evaluate the allocation decision online.
The core of the role allocation mechanism is represented by a modified version of the standard fallback node, named \texttt{Cost-Based Fallback}. Differently from a standard fallback, where children are ticked according to a fixed order, the proposed node dynamically reorders its $N$ children at every tick according to a varying cost vector $\boldsymbol{c} \in \mathbb{R}^N$. 
Each child of the \texttt{Cost-Based Fallback} is an action node $A_i$, with $i \in \{1, \dots, N\}$, triggering the execution of a specific activity included in the task model.

The cost vector $\boldsymbol{c}$ is obtained by element-wise multiplication of $M$ HRC-related cost vectors $\{\boldsymbol{c}^{m}\}_{m=1}^{M}$. Accordingly, the cost associated with activity $A_i$ is computed as:
\[
c_i = \prod_{m=1}^{M} c_i^{m}, \qquad i \in \{1, \dots, N\}
\]
We chose to combine the individual cost terms through multiplication rather than a weighted sum because, besides avoiding the non-trivial problem of weight tuning, it yields a more interpretable decision process, as each factor has a clear and direct effect on the final cost. Moreover, the multiplicative formulation enables hard constraints: if any cost component $c_i^m$ is zero, the overall cost $c_i$ becomes zero, effectively preventing $A_i$ execution by the robot. 
Specifically, at each iteration, the \texttt{Cost-Based Fallback} children are sorted in ascending order of cost so that the activity associated with the lowest positive cost is selected first, while activities associated with zero cost are moved to the end of the list.

The following subsection details the computation of the adopted $\{\boldsymbol{c}^{m}\}_{m=1}^{M}$. However, the proposed BT structure remains agnostic to the specific cost formulation, also allowing additional factors to be seamlessly integrated.

\vspace{-0.3cm}
\subsection{Costs Computation}
\label{ssec:costs_computation}
Some costs among $\{\boldsymbol{c}^{m}\}_{m=1}^{M}$ encode the availability of a specific activity at the current time (those computed by the red blocks in Fig. \ref{fig:role_alloc_framework}), while others regulate activity priority (returned by the yellow blocks). All the proposed cost-computation modules rely on an HRC task model. 

\subsubsection{Task Model}
The HRC task model is represented as a JSON file describing the collaborative task at the activity level. Each activity is characterized by a unique identifier, a textual description, the involved objects, the activities that must precede it, its feasibility for the robot, and a reference representation used for activity recognition and correctness assessment.
An example of a task model is reported below:

{\scriptsize
\begin{verbatim}
{
  "ID": 1,
  "description": "open the drawer",
  "objects involved": ["drawer"],
  "must follow": [],
  "robot feasible": false,
  "reference": R1
},
{
  "ID": 2,
  "description": "place the screwdriver inside the drawer",
  "objects involved": ["screwdriver", "drawer"],
  "must follow": [1],
  "robot feasible": true,
  "reference": R2
},
{
  "ID": 3,
  "description": "place the wrench inside the drawer",
  "objects involved": ["wrench", "drawer"],
  "must follow": [1],
  "robot feasible": true,
  "reference": R3
}
\end{verbatim}
}
where $A_1$ is deemed not feasible for the robot, while $A_2$ and $A_3$ depend on $A_1$ through the \texttt{must follow} field. Please note that, in this work, dependencies are only enabling: completing an activity can unlock new activities but cannot invalidate alternative ones. For instance, if an additional activity $A_4$ ``close the drawer'' were executed, it would require temporarily disabling $A_2$ and $A_3$ until the drawer is reopened.

\subsubsection{History-Based Costs}
\label{sssec:history}
The history-based cost vectors $\boldsymbol{c}^{h-done}$ and $\boldsymbol{c}^{r-done}$ are used to keep track of the activities already completed by the human and the robot, respectively, preventing their re-allocation.
Both vectors are initialized to all ones. Whenever the completion of activity $A_i$ is detected, the corresponding cost element associated with the executing agent is set to zero. As discussed in Section~\ref{ssec:role_allocator}, this is sufficient to prevent the activity from being selected again by the role allocator.
In particular, $c_i^{r-done}$ is updated as soon as $A_i$ is allocated to the robot. 
Thus:
\[
c_i^{r-done} =
\begin{cases}
1 & \text{if the robot has not executed } A_i \\
0 & \text{if the robot is executing or}  \\
 & \text{has already completed } A_i
\end{cases}
\]

On the other hand, activities autonomously executed by the human must be monitored and recognized. The proposed framework allows integrating different activity-recognition approaches for this purpose. In this work, the \textit{Human Activity Handler} exploits the Shannon entropy-based interaction encoding we proposed in \cite{merlo2025exploiting} to identify the ongoing human activity from video monitoring of the environment, as detailed in Section~\ref{ssec:entropy}. The method recognizes both the completion and the correctness of $A_i$, exploiting the information of the \texttt{reference} field, updating $c_i^{h-done}$ as:
\[
c_i^{h-done} =
\begin{cases}
1 & \text{if the human has not executed } A_i \\
0 & \text{if the human has correctly executed } A_i \\
10^{-3} & \text{if the human incorrectly executes } A_i
\end{cases}
\]
Assigning such a low value to $c_i$ when activity $A_i$ is incorrectly executed strongly prioritizes the robot execution of that activity, allowing the robot to immediately recover from the human partner's error.
At the end of the collaborative task, we expect that for each $A_i$ either $c_i^{h-done}$ or $c_i^{r-done}$ is zero.

\subsubsection{Task-Based Costs}
The \textit{Task Constraints Handler} defines which activities are unlocked for the robot according to the task constraints and the current \textit{collaboration status}. 
It takes as input the task model together with the completion vectors $\boldsymbol{c}^{h-done}$ and $\boldsymbol{c}^{r-done}$. 
For each activity $A_i$, the module verifies robot feasibility and prerequisite activities $\mathcal{A}_i^{prec}$ listed in the \texttt{must follow} field: if $A_i$ is not robot-feasible or at least one activity in $\mathcal{A}_i^{prec}$ has not yet been completed, then $c_i^{task}=0$.
In the previous example, activity $A_2$ (``place the screwdriver inside the drawer'') requires $A_1$ (``open the drawer'') to be completed beforehand. Thus, if both $c_1^{h-done} = c_1^{r-done} = 1$, namely $A_1$ has not been executed by either of the two agents, $A_2$ remains locked and $c_2^{task}=0$. Otherwise, the drawer is assumed to be already open and $c_2^{task}=1$.

Moreover, the module receives the currently manipulated object $o_m$ from the Human Activity Handler. Activities involving $o_m$, as well as activities involving objects interacting with $o_m$ within the task model, are inserted into a set of conflicting activities $\mathcal{A}^{conflict}(o_m)$ and temporarily blocked for robot allocation to avoid conflicts with the ongoing human action.
For example, if $o_m$ is the screwdriver, then $\mathcal{A}^{conflict}(o_m)=\{A_2,A_3\}$. $A_2$ is blocked because it directly involves the screwdriver, whereas $A_3$ is disabled since the human may be placing the screwdriver into the drawer, and assigning the robot to $A_3$ could interfere with that action by simultaneously occupying the drawer.
Thus, for each $A_i$, $c_i^{task}$ is defined as:
\[
c_i^{task} =
\begin{cases}
 & \text{if $A_i$ is not feasible for the robot} \\
0 & \text{or } \mathcal{A}_i^{prec} \text{ are not completed} \\
 & \text{or $A_i \in \mathcal{A}^{conflict}(o_m)$} \\
1 & \text{otherwise}
\end{cases}
\]

\subsubsection{Explicit Preference-Based Cost}
The \textit{Preferences Handler} takes as input the task model together with a natural language spoken request that has been transcribed. Both are given to an LLM along with a guideline prompt explaining how to interpret the semantics of the different JSON fields. The LLM is then asked to process the user request and generate a preference cost vector $\boldsymbol{c}^{pref}$, where:
\[
c_i^{pref} =
\begin{cases}
10^{-2} & \text{if the user wants the robot to execute } A_i \\
10^{2} & \text{if the user wants to execute } A_i \\
1 & \text{if no explicit preference is expressed}
\end{cases}
\]
The cost list returned by the LLM is extracted and converted into a numerical vector processable by the BT. Hallucinations are mitigated by grounding the LLM reasoning on the provided task model and by enforcing a structured output format. Generated costs are internally checked and, in case of invalid outputs (i.e., values different from $10^{-2}, 10^{2}$, or $1$), $\boldsymbol{c}^{pref}$ is initialized to all ones.
This vector highly influences the execution priority among the available activities, prioritizing those associated with $c_i^{pref} = 10^{-2}$ for robot execution.

\subsubsection{Distance-Based Costs}
The \textit{Environmental Features Handler} computes the cost for each activity $A_i$ simply as:
\[
c_i^{dist} = \frac{1}{d_{ho_{m,i}}}
\]
where $d_{ho_{m,i}}$ denotes the distance, expressed in meters, 
between the operator's hand position $\boldsymbol{p}_h$ and the position of the object manipulated in $A_i$ $\boldsymbol{p}_{o_{m,i}}$, provided by the object tracker.
All costs in $\boldsymbol{c}^{dist}$ are subsequently normalized by dividing each $c_i^{dist}$ by the maximum cost computed over the set of activities $\max(c_i^{dist})$. As a result, the normalized costs belong to the range $[0,1]$, where values closer to 1 indicate activities involving objects that are more easily reachable by the human.
The $o_m$ associated with each activity is retrieved by inspecting the task model.
This cost vector has a lower priority with respect to the one encoding the user's explicit preferences. However, when the higher-priority cost vector does not determine a unique best activity for the robot, the one with the lowest $c_i^{dist}$ is chosen.

Note that ergonomic indexes could be integrated at this level.

\vspace{-0.2cm}
\subsection{Entropy-Based Human Activity Identification}
\label{ssec:entropy}

The Human Activity Handler in this framework exploits the Shannon entropy and Mutual Information (MI)-based interaction encoding we introduced in \cite{merlo2025exploiting, merlo2025information}. 
The original method processes offline RGB video demonstrations by extracting hand-object (HO) and object-object (OO) interactions over time. A brief summary on how it works follows.
Starting from tracked $6$D hand and object poses, the average hand-object distance $\overline{d}_{ho}(t^*)$ and the MI $MI_{ho}(t^*)$ are estimated over a sliding temporal window $w$ centered around each instant $t^*$. If $\overline{d}_{ho}(t^*) < d_{ho}^{th}$ and $MI_{ho}(t^*)>0$, the two signals are considered statistically dependent, indicating an ongoing HO interaction. By shifting $w$ over time, a temporal $MI_{ho}(t)$ profile is obtained, whose evolution characterizes the manipulation dynamics.
The method also detects stable OO by analyzing the entropy variation of the relative distance $H(\overline{d}_{oo})$ between the manipulated object $o_m$ and a static background object $o_{bkg}$. In particular, $\overline{d}_{oo}(t^*) < d_{oo}^{th}$ and $\frac{dH(\overline{d}_{oo}(t^*))}{dt}<0$ indicate convergence toward a stable interaction configuration, whereas temporary OO are filtered out.
Each interaction detected while processing frame $k$ (at instant $t^*$) is represented through a scene graph $G_k$. In the simplest HO case, $G_k$ contains two nodes representing the hand $h$ and $o_m$, connected through an edge encoding $MI_{ho}(t^*)$ and their relative pose as a transformation matrix $T_{h \rightarrow o_{m}}$. When an OO is detected, a third node representing $o_{bkg}$ is added, and the edge connecting $o_m$ and $o_{bkg}$ includes $T_{o_m \rightarrow o_{bkg}}$.
Once the whole video is processed, the generated graph stream $\{G_k\}$ is segmented whenever the topology changes, generating temporal segments called Interaction Units (IUs). Consecutive IUs sharing the same $o_m$ are grouped into higher-level activities, naturally separated by idle-hand phases.
Each IU is compactly represented through a subset of $\{G_k\}$ extracted at the local minima of $MI_{ho}$, since low variability in hand-object information sharing typically corresponds to temporally stable interaction configurations. Consequently, each activity $A_i$ is represented as an ordered sequence of representative graphs
$\mathcal{G}_i = \{G^{repr}_{i,j}\}_{j=1}^{J_i}$
where $J_i$ is the number of key interaction configurations happening in $A_i$.

In such a module, the original offline method is adapted for generating graphs online, analyzing the RGB data of the human execution. Since future samples are unavailable online, the temporal window $w$ can no longer be centered around the current instant. Therefore, $\overline{d}_{ho}(t)$ and $MI_{ho}(t)$ are estimated only from past observations in the interval $\left[t^*-w,\, t^*\right]$. Activity segmentation is also performed online.
The first output of this module is $o_m$, sent to the Task Constraints Handler.
To robustly estimate it, the module waits until a buffer of $n$ consecutive $G_k$ sharing the same HO is collected.
The second output is the identification of the completed human activity and its correctness. To this end, the \texttt{reference} field in the task model encodes the expected sequence of representative graphs $\mathcal{G}_i^{ref}$ associated with the correct execution of $A_i$. In our example, for $A_2$ two IUs are expected: IU$_1$ involving the hand and the screwdriver, IU$_2$ encodes an HO+OO involving the hand, the screwdriver, and the drawer during insertion. Then,  $\mathcal{G}_2^{ref} = \{G_{2,1}^{repr}, G_{2,2}^{repr}\}$, encoding the relative pose between the hand and the screwdriver at the beginning of the manipulation and the relative pose between the screwdriver and the drawer at its end.
Whenever an online activity segment terminates, the corresponding observed representation
$\mathcal{G}^{obs}$
is extracted and compared against all the reference activity representations
$\{\mathcal{G}_i^{ref}\}_{i=1}^{N}$
stored in the task model.
The comparison is performed at three levels:
(i) number of key interaction configurations,
(ii) graph topology,
and (iii) relative object layout (see Algorithm \ref{alg:activity_identification}).
In particular, the relative OO configurations are compared through the transformation matrices $T^{obs}_{o_m \rightarrow o_{bkg}}$ and $T^{ref}_{o_m \rightarrow o_{bkg}}$. The observed and reference layouts are considered compatible if
\begin{equation}
\label{eq:TTinv}
\Delta T = T^{obs}_{o_m \rightarrow o_{bkg}}
\left(T^{ref}_{o_m \rightarrow o_{bkg}}\right)^{-1}
\approx I   
\end{equation}
according to predefined distance thresholds.
For example, once $A_3$ is completed, thus the wrench is placed inside the drawer, the current history $\mathcal{H}$ is segmented and the representative graphs extracted from the HO and HO+OO phases compose $\mathcal{G}^{obs}$. The comparison with $\mathcal{G}_1^{ref}$ is rejected due to representative graphs topology describing only an HO phase (between hand and the drawer), while $\mathcal{G}_2^{ref}$ is discarded because the manipulated object $o_m$ differs. Finally, $\mathcal{G}^{obs}$ is compared against $\mathcal{G}_3^{ref}$ through the OO layouts: if
$\Delta T \approx I$ with $o_m$ the wrench and $o_{bkg}$ the drawer, the activity is recognized as correctly completed; otherwise, an anomaly is reported for $A_3$.

\begin{algorithm}[t]
\caption{Entropy-based human activity identification}
\label{alg:activity_identification}
\begin{algorithmic}[1]
\algrenewcommand\algorithmiccomment[1]{\hfill{\scriptsize$\triangleright$ #1}} \small
\Require Online scene graph stream $\{G_k\}$
\Require Activity references $\{\mathcal{G}_i^{ref}\}_{i=1}^{N}$ \Comment{From task model}

\State $\mathcal{H} \gets \emptyset$ \Comment{Initialize current graph history}
\State $\boldsymbol{c}^{h\text{-}done} \gets \boldsymbol{1}$ \Comment{Initialize $c^{h-done}$ costs}

\For{each incoming $G_k$}
    \If{an idle phase is detected}
        \State Segment the current history $\mathcal{H}$
        \State Extract $\mathcal{G}^{obs}$ from $\mathcal{H}$ \Comment{Local minima of $MI_{ho}$ are used}
        
        \For{each $i$}
            \State Retrieve $\mathcal{G}_i^{ref}$
            
            \If{$|\mathcal{G}^{obs}| \neq |\mathcal{G}_i^{ref}|$} \Comment{Check number of key configurations}
                \State \textbf{continue}
            \EndIf
        
            \State $\mathcal{D}_T \gets \emptyset$ \Comment{Initialize OO layout errors}

            \For{each $({G}_j^{obs},{G}_{i,j}^{ref})$} 
                \If{\texttt{nodes}$({G}_j^{obs}) \neq \texttt{nodes}({G}_{i,j}^{ref})$} 
                    \State \textbf{break} \Comment{Topology mismatch} 
                \EndIf
            
                \If{${G}_j^{obs}$ and ${G}_{i,j}^{ref}$ include an OO interaction} 
                    \State Compute $\Delta T$ using Eq.~\ref{eq:TTinv}
                    \State Append $\Delta T$ to $\mathcal{D}_T$
                \EndIf
            \EndFor
            
            \If{all $\Delta T \in \mathcal{D}_T$ satisfy $\Delta T \approx I$}
                \State $c_i^{h\text{-}done} \gets 0$ \Comment{$A_i$ correctly done}
            \Else
                \State $c_i^{h\text{-}done} \gets 10^{-3}$ \Comment{$A_i$ anomalously performed}
            \EndIf
            \State \textbf{return} updated $\boldsymbol{c}^{h\text{-}done}$
            \State $\mathcal{H} \gets \emptyset$
            
        \EndFor 
        
    \EndIf
    
    \State Append $G_k$ to $\mathcal{H}$
\EndFor
\end{algorithmic}
\end{algorithm}

\begin{figure}[t]
    \centering
    \includegraphics[width=\linewidth]{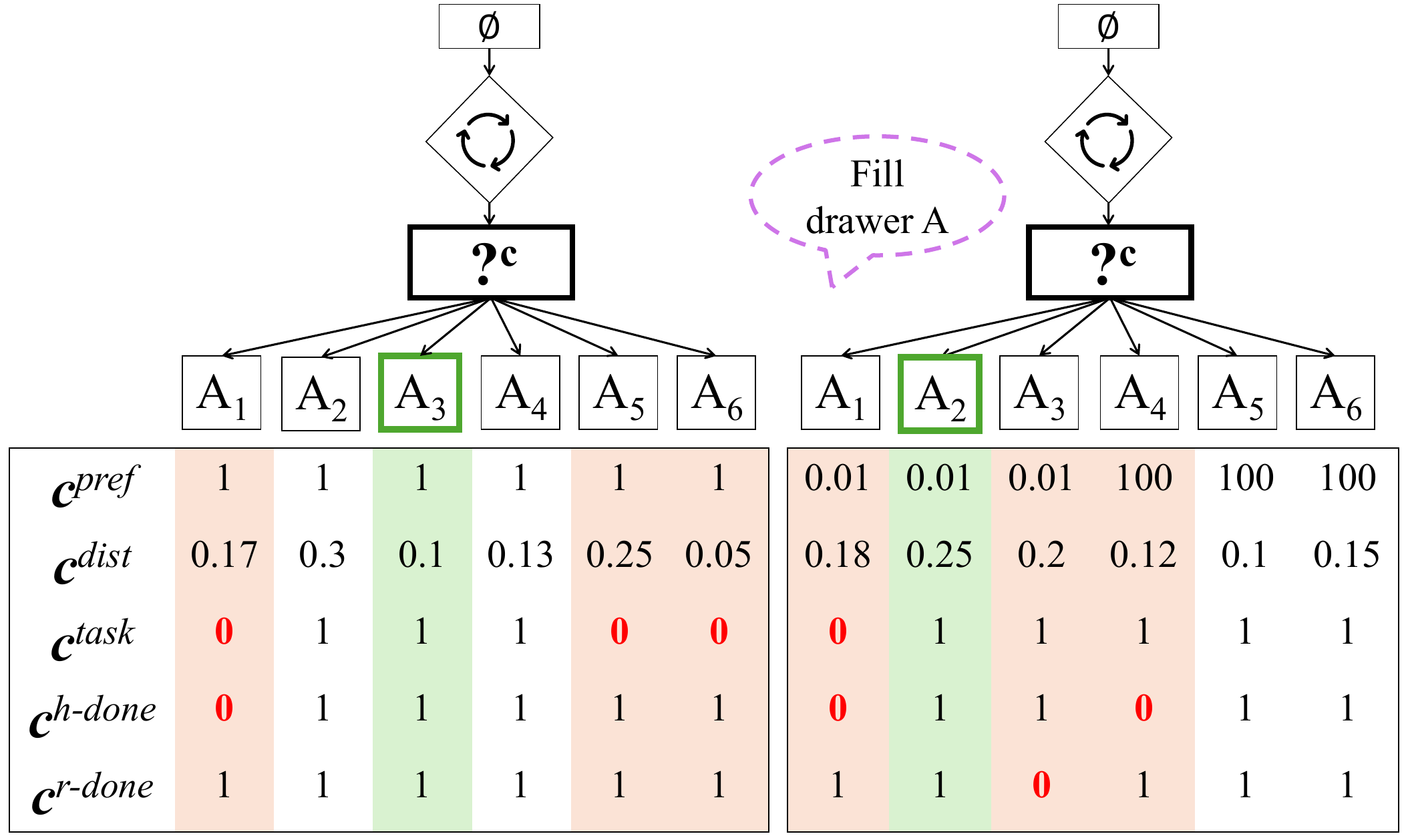}
    \vspace{-0.6cm}
    \caption{Example of how the cost vectors are combined to select the activity to be allocated to the robot. To enable the Fallback logic to operate correctly, the activity highlighted in green is placed at the top of the children node list, whereas the activities highlighted in red are placed at the bottom.}    
    \vspace{-0.5cm}
    \label{fig:explain_alloc}
\end{figure}

\vspace{-0.3cm}
\subsection{Overview of the Allocation Strategy}
\label{ssec:example_alloc}
To conclude this section, we illustrate the allocation process through a simple example. Starting from the task model described in Section \ref{ssec:costs_computation}, let us consider a second drawer, B, that will host the hammer and the saw, in addition to drawer A that will contain the screwdriver and the wrench. This extension adds three activities: $A_4$ (``open drawer B''), $A_5$ (``place the hammer inside drawer B''), and $A_6$ (``place the saw inside drawer B''), with $A_5$ and $A_6$ depending on $A_4$.
The robot can execute all activities except $A_1$. Fig. \ref{fig:explain_alloc} shows the BT and the cost vectors responsible for ordering the \texttt{Cost-Based Fallback} children. In the situation depicted on the left, only $A_1$ has already been completed by the human, thus $c_1^{h-done}=0$. Since $A_1$ is unfeasible for the robot, also $c_1^{task}=0$. Moreover, $c_5^{task}=c_6^{task}=0$ because $A_4$ has not been executed yet. As no user preference has been expressed, all $c_i^{pref}=1$.
Activities with at least one zero cost are not allocatable, leaving $A_2$, $A_3$, and $A_4$ as candidates. Based on the normalized distance costs $\boldsymbol{c}^{dist}$, $A_3$ is assigned to the robot since it involves the object farthest from the human, namely the wrench.
Then, let us assume that while the robot executes $A_3$, the human completes $A_4$ and requests: \textit{``Fill drawer A''}. The updated costs are shown in Fig. \ref{fig:explain_alloc} (right). The preference vector assigns $10^{-2}$ to activities involving drawer A and $10^2$ to those involving drawer B. Since $A_3$ and $A_4$ have been completed, they cannot be reassigned, while the execution of $A_4$ unlocked $A_5$ and $A_6$ ($c_5^{task}=c_6^{task}=1$). Although $A_5$ involves the object farthest from the user, the explicit preference forces $A_2$ to be selected for robot execution.

\vspace{-0.2cm}
\section{Experiments}
\label{sec:experiments}
The experimental campaign aimed at evaluating the performance of 
the proposed role allocation framework from both a technical and a user-centered perspective. 
First, a pilot study was conducted to demonstrate all the functionalities of the framework, including activity recognition, preference handling, and anomaly management\footnote{The multimedia attachment related to this experiment can be found 
online at \href{https://youtu.be/JDbyG3rafLQ}{https://youtu.be/JDbyG3rafLQ}.}. Subsequently, some participants were involved in a user study in which they collaborated with the robot using the proposed allocation strategy to evaluate the impact of the full framework on the \textit{intuitiveness} and \textit{human agency}, rather than test each isolated component. 

\vspace{-0.2cm}
\subsection{Role Allocation Framework Validation (Exp 1)}
\label{ssec:expesetup-pilot}
\subsubsection{Experimental Setup}
\begin{figure}[t]
    \centering
    \includegraphics[width=0.8\linewidth]{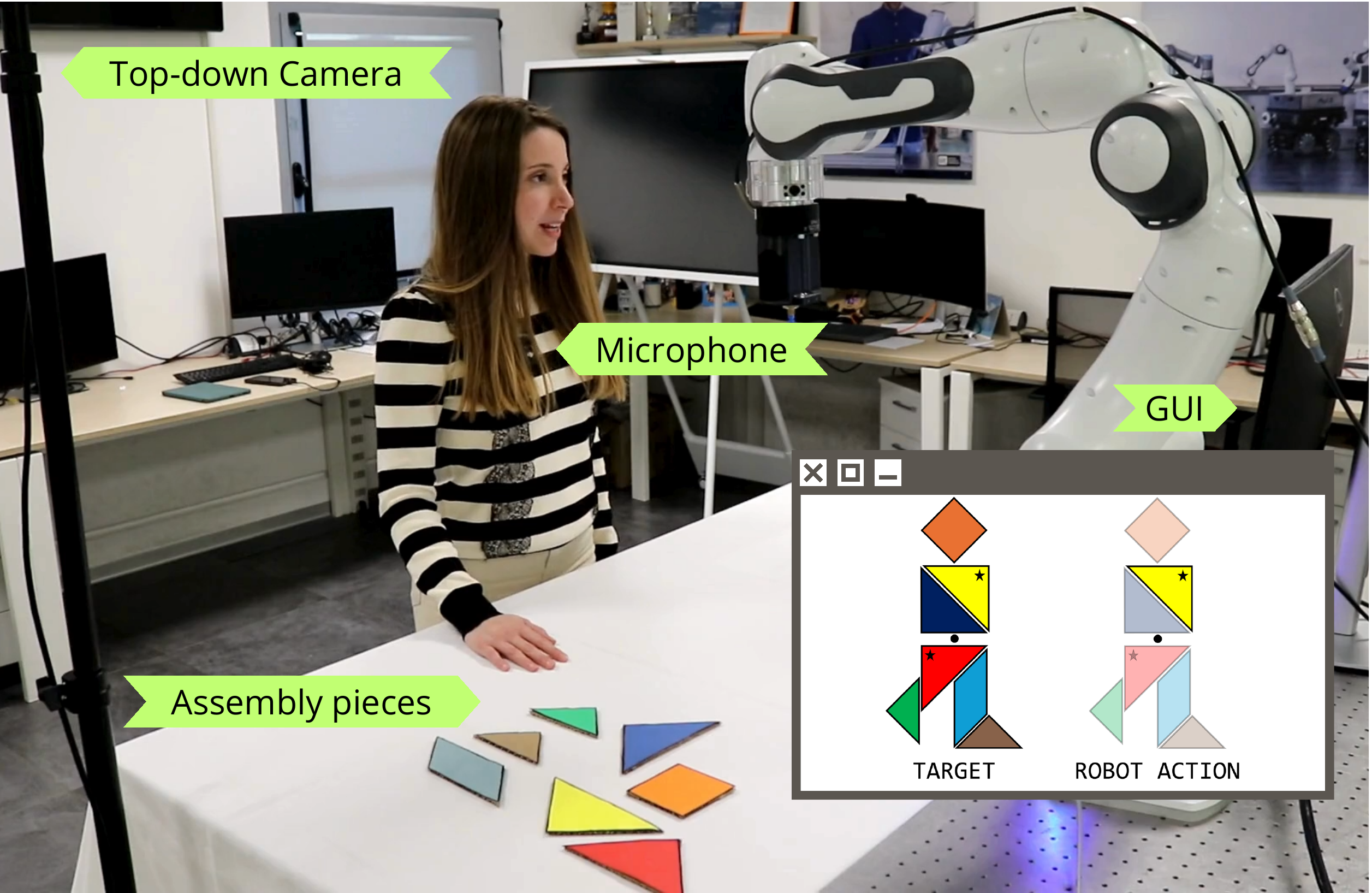}
    \vspace{-0.3cm}
    \caption{Experimental setup used for testing the proposed framework.}    
    \vspace{-0.5cm}
    \label{fig:expe-setup}
\end{figure}

The experimental setup is shown in Fig.~\ref{fig:expe-setup}. A human participant and a fixed-base Franka Emika Panda manipulator, equipped with a Robotiq EPick vacuum gripper, were positioned on opposite sides of a shared workspace. A RealSense D435i camera, mounted above the table on an L-shaped support, provided a top-down view of the interaction area.
Hand and object tracking was performed from the RGB stream using SAM 2 \cite{ravi2025sam}. To enable online operation, we implemented a stateful tracker that processes the live camera stream, initializes targets from user-provided prompts in the first frame, and propagates segmentation masks over subsequent frames. This provides temporally consistent tracking without requiring object classification. The image-plane centroid of each binary mask was used as the object position. Under partial occlusions, positions were updated only when the visible mask area exceeded a predefined fraction of its reference area; otherwise, the last valid estimate was retained. Tracking seven objects and one hand, the complete perception pipeline operated at approximately 7 Hz.
The participant wore a microphone to communicate explicit preferences. 
The GPT-5.1 model was used to process them. 
A monitor placed next to the robot displayed a graphical user interface (GUI) reporting both the target assembly and the activity currently assigned to the robot. While the participant was free to perform any activity, the GUI informed them about the robot allocation decision, reducing the risk of conflicts during task execution.
The task model structure was handcrafted, whereas both the activity references $\{\mathcal{G}_i^{ref}\}_{i=1}^{N}$ and the robot primitives executed by the children of the \texttt{Cost-Based Fallback} node were automatically generated from a human demonstration using the method proposed in \cite{merlo2025exploiting}. 
The $o_{bkg}$ was defined as a point on the table corresponding to the center of the final assembly. The distance thresholds were set to $\overline{d}_{ho}^{th}=0.15$ m and $\overline{d}_{oo}^{th}=0.20$ m. The temporal window was set to $w=30$ samples, corresponding to approximately $1$ s of observations at the acquisition frequency of $30$ Hz.
The cost vectors are updated at a rate of $100$ Hz.

\subsubsection{Experimental Protocol}
The collaborative task consisted of assembling a \href{https://en.wikipedia.org/wiki/Tangram}{Tangram} figure composed of seven colored pieces (five triangles, one square, and one parallelogram). The target figure was a human-shaped silhouette, shown in the GUI of Fig.~\ref{fig:expe-setup}. Each activity consisted of moving one piece from the picking area to its target position, yielding an HRC task model with $N=7$ robot-feasible activities.
The figure was divided into two subassemblies (upper and lower body), each anchored by a pivotal piece (yellow and red, respectively) that had to be placed first. The remaining pieces could only be positioned adjacent to already assembled ones, resulting in the precedence constraints encoded in the \texttt{must follow} field. The task model also associated each piece with its shape, color, and semantic role (e.g., head, torso, leg, or foot).
An expert participant evaluated the framework. She was instructed to express one preference before starting the task and another during execution, comply with the task constraints, and manipulate the pieces using only her right hand, which was tracked by the system.
We selected this task because it captures key characteristics of collaborative assembly: structural constraints that induce a partial order while allowing parallel execution of independent activities. This resembles many industrial assembly processes, where multiple valid execution sequences exist as long as the final product is correctly assembled. Moreover, the diversity of the pieces in terms of color, shape, and semantic role enables a wide range of user preferences, making the task well suited for evaluating the proposed Preferences Handler.
The experiment was designed to demonstrate the framework's main capabilities.

\subsection{Multi-subject Experience Evaluation (Exp 2)}
\begin{figure}[b]
    \centering
    \vspace{-0.5cm}
    \includegraphics[width=0.45\linewidth]{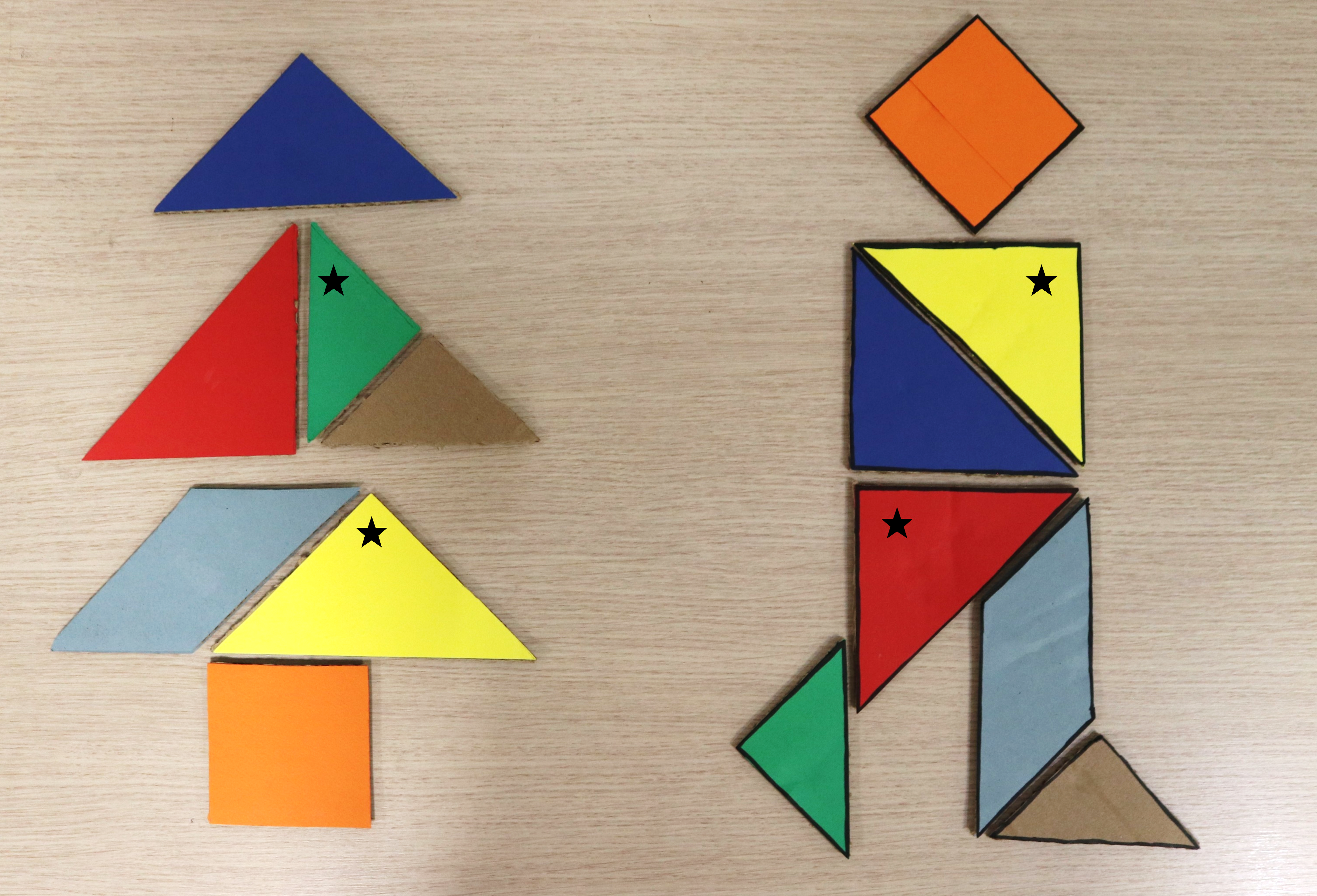}
    \caption{Final assembly in the multi-subject experiment. The stars on top indicate the pivotal pieces.}
    \label{fig:final_assembly}
\end{figure}

\subsubsection{Experimental Setup}

The experimental setup remained unchanged (Fig.~\ref{fig:expe-setup}), but the task was extended to better evaluate the proposed role allocation strategy in a more realistic collaborative scenario and assess its effects on user experience and task performance. Participants assembled two Tangram figures side by side: the human-shaped silhouette used in the pilot study and a tree silhouette (Fig.~\ref{fig:final_assembly}), resulting in $N=14$ activities. Pieces outlined in black formed the human figure, whereas the remaining pieces formed the tree. The stars indicate the pivotal pieces.
Due to the increased number of tracked objects, the SAM-based perception system could no longer provide a sufficiently high update rate for computing the cost vectors. Consequently, only the participant's hand was tracked online to compute $\boldsymbol{c}^{dist}$, while the initial 6D poses of the Tangram pieces were predefined. Human activity recognition was performed manually by the experimenter.

\subsubsection{Experimental Protocol}
We recruited 14 participants (9 males and 5 females, age $29.4 \pm 2.6$ years) from students and staff at IIT\footnote{Experiments were carried out at the HRII Lab., in accordance with the Helsinki Declaration, and the protocol was approved by the Comitato etico territoriale Regione Liguria (Protocol IIT\_HRII\_002).}.
Each participant completed the collaborative task twice under two conditions presented in randomized order: (A) the proposed preference-based role allocation strategy and (B) a baseline approach inspired by existing HRC methods, in which both agents receive explicit task assignments.
In condition A, participants were free to contribute at any time, while activities were dynamically allocated to the robot based on the current \textit{collaboration status}. They were instructed to express at least one natural language preference before starting the task and could optionally provide another during execution.
In condition B, activities were simultaneously assigned to both the human and the robot through the GUI, as done in \cite{merlo2023ergonomic}. Participants were required to perform only their assigned activities. The allocation strategy relied solely on $\boldsymbol{c}^{task}$ and $\boldsymbol{c}^{dist}$, assigning each agent a feasible activity that satisfied the task constraints while favoring the closest available piece for the human.

\begin{figure*} [t]
    \centering
    \includegraphics[width=0.9\linewidth]{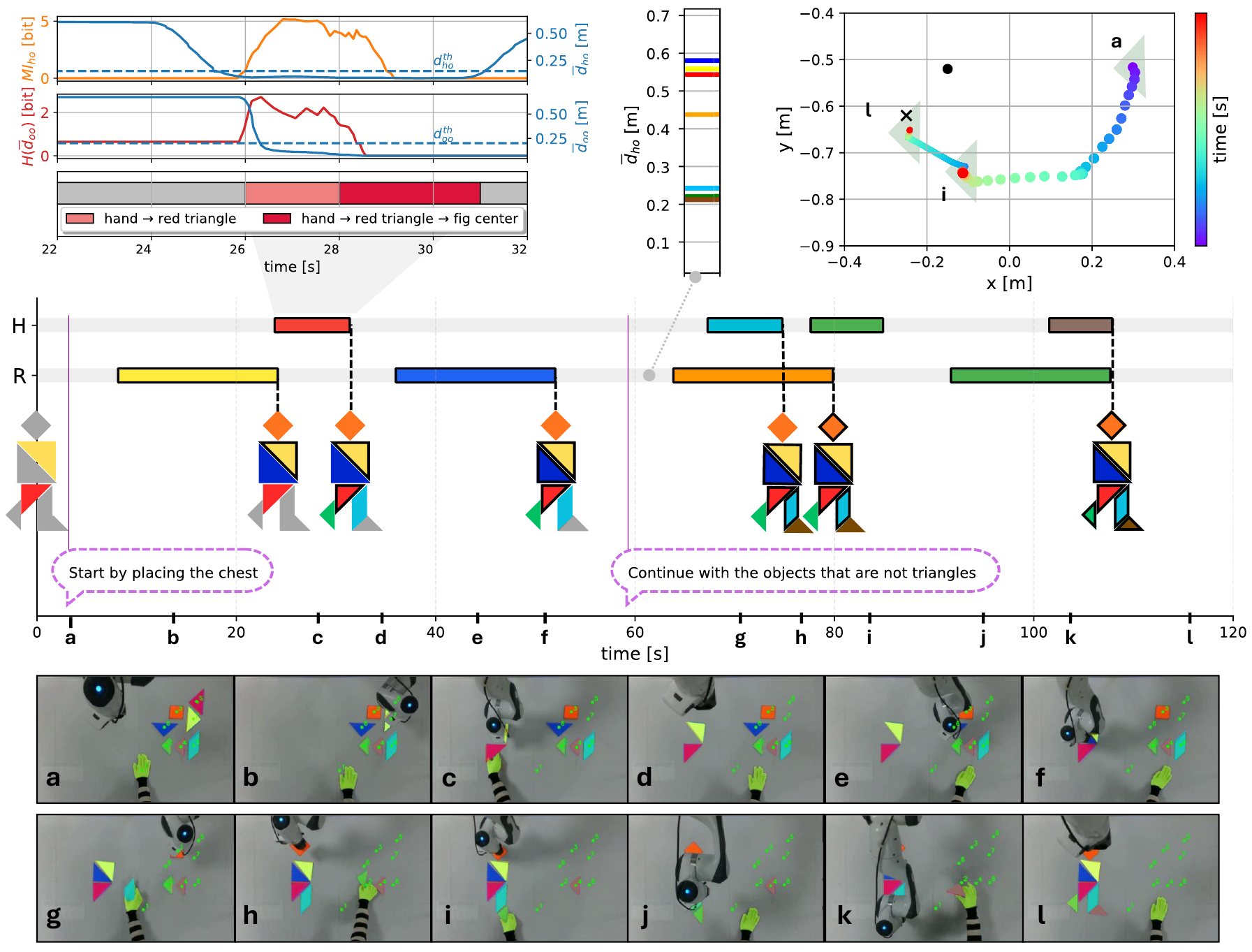}
    \vspace{-0.2cm}
    \caption{Pilot experiment results demonstrating the main functionalities of the proposed framework. The central timeline reports the activities executed by the human (H) and the robot (R), together with the evolution of the task status, visualized through the Tangram silhouette and the color changes of its pieces, and the explicitly asked preferences shown in the balloons. (Upper-left) Online computation of the entropy-based interaction measures used for activity segmentation and recognition. (Upper-middle) Hand-object distance estimates used to compute the distance-based costs. (Upper-right) Detection and correction of an incorrectly executed activity. (Bottom) Snapshots acquired from the overhead camera.}
    \vspace{-0.4cm}
    \label{fig:pilot}
\end{figure*}

\subsubsection{Metrics and statistics}
After each condition, participants completed a custom questionnaire and the NASA-TLX. The custom questionnaire evaluated the two primary objectives of the proposed framework: \textit{intuitiveness} and \textit{human agency}. Specifically, participants rated the following statements on a 7-point Likert scale:

\textit{In1:} It was clear how the robot selected the action to take.

\textit{In2:} The robot's next action was predictable.

\textit{Ag1:} I felt forced by the system to perform a specific action.

\textit{Ag2:} I enjoyed influencing how roles were assigned.

Participants also indicated which factors they believed influenced the role allocation under each condition. Available options included \textit{hand-object distance},\textit{ voice commands}, and \textit{task constraints}. To assess the intuitiveness of the allocation process, they could alternatively select \textit{random criteria}, \textit{I did not pay attention}, or \textit{I could not understand}. An additional statement, \textit{The robot correctly interpreted my requests and acted consistently}, evaluated satisfaction with the handling of explicit preferences.
Finally, participants reported their preferred condition and briefly justified their choice. Differences between conditions were analyzed using two-sided Wilcoxon signed-rank tests for paired observations. The same analysis was applied to the NASA-TLX scores to compare perceived workload.
Two objective performance metrics were also collected: task execution time (the time required to complete the assembly) and human idle time (the cumulative duration during which the participant's hand remained stationary), the latter serving as an indicator of passive waiting during the collaboration.

\vspace{-0.3cm}
\section{Results}
\subsection{Role Allocation Framework Validation (Exp 1)}
Fig.~\ref{fig:pilot} reports the timeline of the pilot experiment. The lower part of the figure shows snapshots from the top-down camera with SAM-generated segmentation masks overlaid on the tracked hand and objects. Numbered markers indicate the object initialization performed before the task.
The horizontal bands labeled \textit{H} and \textit{R} represent the human and robot activity histories. Colored segments denote executed activities, where the color identifies the manipulated Tangram piece and the segment length corresponds to its execution time. Robot activity durations are fixed, whereas human durations are estimated online by the entropy-based Human Activity Handler.
The Tangram silhouette above the timeline summarizes the task status: grey pieces denote unavailable activities ($c_i^{task}=0$), colored pieces available activities ($c_i^{task}=1$), and black-outlined pieces completed activities ($c_i^{h-done}=0$ or $c_i^{r-done}=0$).
Initially, only the two pivotal pieces were available. Following the user's request \textit{``Start by placing the chest''}, the Preferences Handler prioritized the yellow and blue torso pieces. Since only the yellow piece satisfied the task constraints, it was assigned to the robot. Once placed, the blue torso and head became available, allowing the robot to continue with the blue torso while the user independently assembled the lower body.
The upper-left plot of Fig.~\ref{fig:pilot} illustrates the online interaction measures used to recognize the placement of the red leg. As the user approached and grasped the piece, the average hand-object distance $\overline{d}_{ho}$ decreased while $MI_{ho}$ increased, indicating an HO. As the piece approached the figure center, the entropy of the object-object distance $H(\overline{d}_{oo})$ decreased, indicating an OO. These graph topology changes segmented the activity into two IUs, i.e., an HO phase followed by an HO+OO phase. The resulting $\mathcal{G}^{obs}$ matched $\mathcal{G}^{ref}_{red}$, marking the activity as completed and unlocking the remaining lower-body activities.
Later, the second explicit request \textit{``Continue with the objects that are not triangles''} prioritized the cyan parallelogram and orange square. The final allocation was determined by the distance cost, assigning the orange square to the robot because it was farther from the user (see the $\overline{d}_{ho}$ signals in the top-middle plot of Fig.~\ref{fig:pilot}).
Finally, the user placed the green foot in an incorrect position. Although $\mathcal{G}^{obs}$ matched the expected activity topology, the observed OO layout did not satisfy the reference configuration. Consequently, the activity was flagged as anomalous ($c_{green}^{h-done}=10^{-3}$), triggering immediate robot intervention. The robot retrieved the misplaced piece and relocated it correctly before the user completed the assembly by placing the final remaining piece, i.e., the brown foot.
The upper-right plot of Fig.~\ref{fig:pilot} shows the relocation of the green piece. Trajectory \textit{a--i} corresponds to the incorrect human placement, \textit{i--l} the robot's corrective motion toward the target \textit{x}.

\subsection{Multi-subject Experience Evaluation (Exp 2)}

\begin{figure*} [t]
    \centering
    \includegraphics[width=\linewidth]{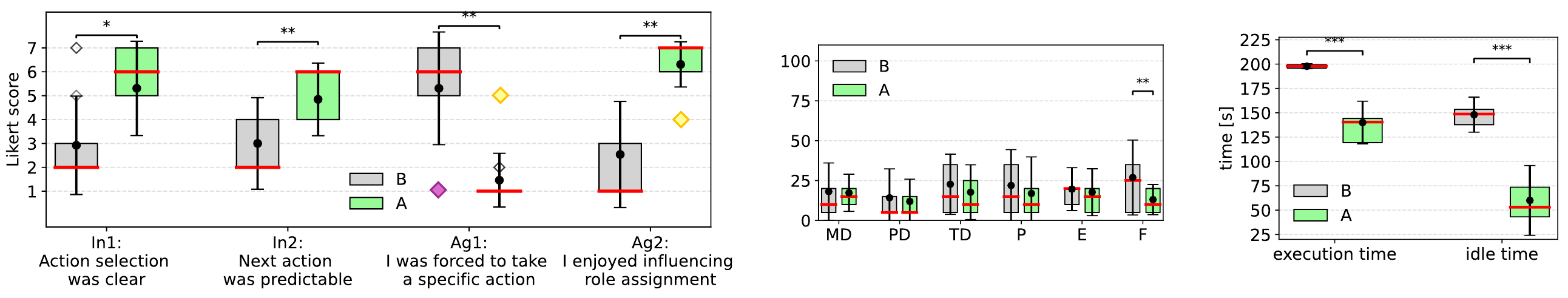}
    \vspace{-0.8cm}
    \caption{(Left) Custom questionnaire items rated by the participants for conditions A and B. Boxes indicate the interquartile range, the red line the median, and black dots with error bars the mean $\pm$ SD. Diamond markers denote outliers. Asterisks indicate significant differences based on paired two-sided Wilcoxon signed-rank tests ($^* p<0.05$, $^{**} p<0.01$, $^{***} p<0.001$). (Middle) NASA-TLX results comparing A and B: Mental Demand (MD), Physical Demand (PD), Temporal Demand (TD), Performance (P), Effort (E), and Frustration (F). (Right) Execution and idle time in the two conditions.}
    \label{fig:questionnaires_results}
    \vspace{-0.5cm}
\end{figure*}

The questionnaire results are reported in Fig.~\ref{fig:questionnaires_results}. The left panel presents the scores of the custom questionnaire. Condition A yielded significantly higher ratings for both the clarity of the robot's action selection (In1, $p<0.05$) and the predictability of its behavior (In2, $p<0.01$).
Consistently, all participants in condition A correctly identified voice commands as one of the factors influencing the robot allocation. This agrees with the fact that the LLM correctly interpreted 100\% of the user requests and generated appropriate allocation priorities. In contrast, under condition B only 31\% of participants correctly identified task constraints and hand-object distance as allocation factors. Among the remaining participants, 38\% reported they could not understand the allocation rationale, 30\% stated they had not paid attention to it, and 1\% believed the allocation was random.
The agency-related questions revealed an even larger difference between conditions. For both Ag1 and Ag2, participants felt significantly less constrained and more involved under condition A. Nearly all participants reported feeling forced to assemble specific pieces in condition B, whereas they perceived greater freedom and engagement in condition A.
Two exceptions were observed. Participant 3 (purple) gave similar ratings in both conditions and, as reported in Table~\ref{tab:preferred_condition}, appreciated both collaboration modalities equally. Participant 10 (yellow) was the only participant who preferred condition B, explaining that they trusted the robot's allocation decisions because they believed it was optimizing the collaboration, despite not understanding its rationale.
The NASA-TLX results are shown in the central panel of Fig.~\ref{fig:questionnaires_results}. The two conditions produced comparable scores across most dimensions. Condition A resulted in a slightly higher Mental Demand (MD), likely reflecting the more active role of the participant. However, the difference was not significant ($p=0.52$), and a moderate increase in mental engagement can be considered desirable in collaborative tasks. The only significant difference was observed for Frustration (F), which was substantially lower in condition A ($p<0.01$).
The right panel of Fig.~\ref{fig:questionnaires_results} reports the task execution time and human idle time. Both metrics differed significantly between conditions ($p<0.001$). In condition B, execution times were nearly constant because the collaboration pace was dictated by the robot and the task ended after a fixed number of robot actions. In contrast, condition A allowed participants to drive the collaboration, resulting in shorter execution times and lower idle times, indicating reduced passive waiting and greater user involvement.

\begin{table}[t]
\centering
\caption{Preferred condition and subjects' motivations.}
\label{tab:preferred_condition}
\scriptsize
\begin{tabularx}{\columnwidth}{|c|c|c|X|}
\hline
\textbf{Subj.} & \textbf{Ord.} & \textbf{Pref.} & \textbf{Why} \\
\hline
1  & BA & A  & I was in charge and the robot followed what I said. \\
2  & AB & A  & More enjoyable task division and piece selection. \\
\rowcolor{violet!25}
3  & BA & -- & -- \\
4  & AB & A  & It was faster to finish the task. \\
5  & BA & A  & I was in control. \\
6  & BA & A  & Better robot understanding and collaboration. \\
7  & AB & A  & Faster and easier to know the actions I should take. \\
8  & BA & A  & It seemed more collaborative. \\
9  & AB & A  & I could control robot actions. \\
\rowcolor{yellow!25}
10 & AB & B  & The robot appeared to follow a strategy; I trusted it. \\
11 & BA & A  & I defined the robot work area, improving predictability. \\
12 & BA & A  & I liked directing the robot. \\
13 & AB & A  & I prefer to decide what to do. \\
14 & AB & A  & It was funnier and faster. \\
\hline
\end{tabularx}
\vspace{-0.5cm}
\end{table}

\begin{figure} [t]
    \centering
    \includegraphics[
    width=0.9\linewidth,
    trim=0.8cm 0cm 0cm 0cm,
    clip
]{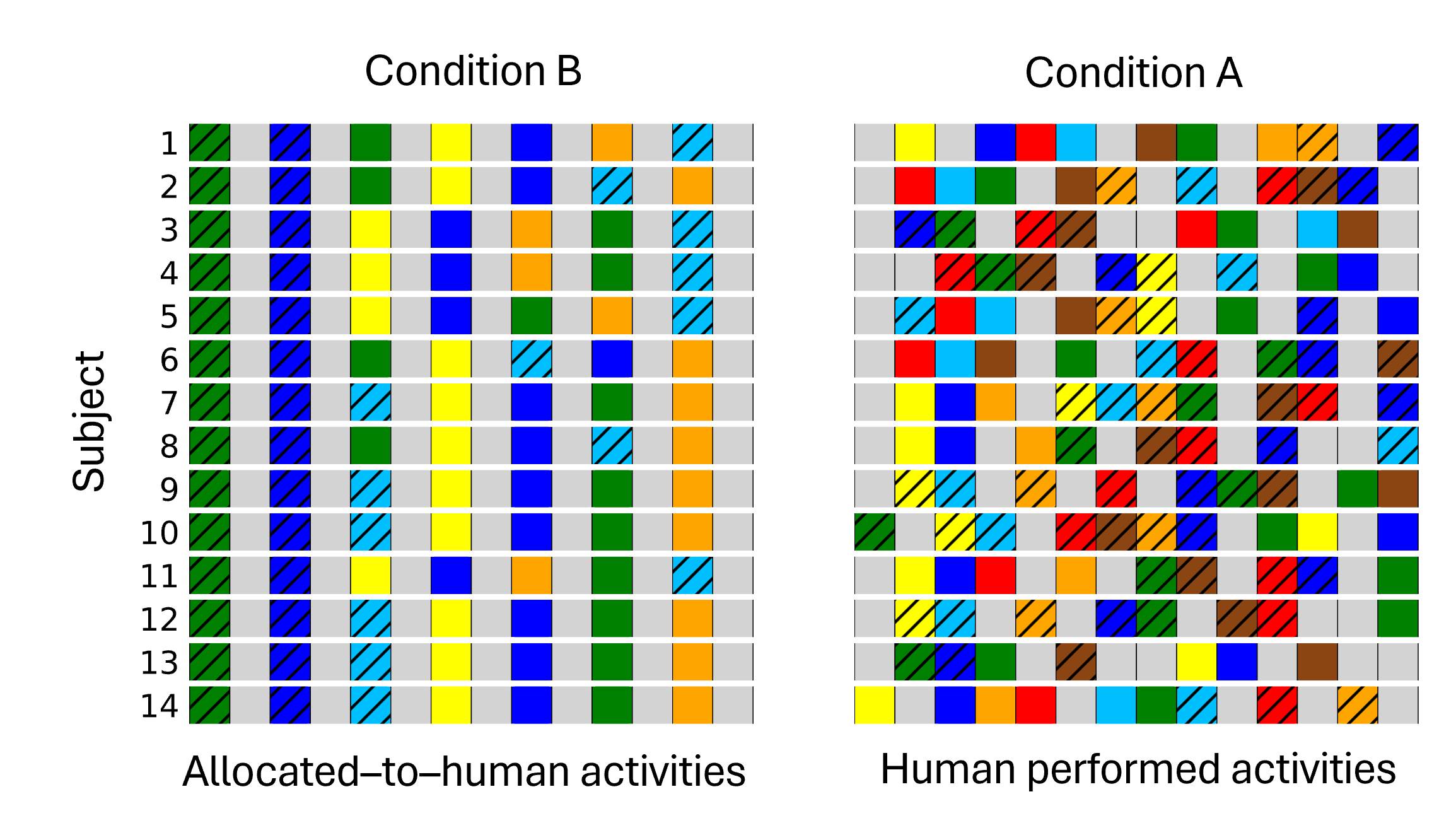}
    \vspace{-0.4cm}
    \caption{Activities executed across participants under the two experimental conditions, either assigned (condition B) or chosen (condition A). Each row represents one participant, and each square corresponds to one of the 14 task activities. Grey squares denote robot-completed activities, while each colored square is the exact human-moved piece (solid: man silhouette; hatched: tree silhouette). Squares are ordered by activity start time. The figure highlights the greater variability of human behavior in condition A compared to condition B.}
    \vspace{-0.5cm}
    \label{fig:alloc_var}
\end{figure}

Another notable result is reported in Fig.~\ref{fig:alloc_var}, which illustrates the effect of human agency on role allocation. The figure shows the activities executed by the human under the two experimental conditions: assigned by the system in condition B and freely selected in condition A. Each row comprises 14 squares, corresponding to the activities in the task model. Grey squares denote robot-executed activities, whereas colored squares indicate Tangram pieces assembled by the human. Solid and hatched patterns distinguish pieces belonging to the human and tree silhouettes, respectively. Activities are ordered by their start time, as both agents operated in parallel throughout the task.
Condition B exhibits little variability across participants. Most subjects were assigned nearly identical activities in a similar order: the first two activities were identical for all participants, while only minor differences appeared thereafter. This stems from the allocation strategy relying primarily on $c^{dist}$. Since the initial piece locations were fixed and participants generally kept their hands within the picking area, the resulting allocations were highly consistent. Moreover, because activities were assigned simultaneously to both agents, the workload was always evenly split, with seven activities allocated to each.
In contrast, condition A shows substantial variability. Participants differed in both activity selection and execution order, completing an average of $8.7$ activities. No two execution patterns were identical.
The explicit preferences expressed by the participants were grouped into categories. Most requests combined spatial and global semantic information, referring to an entire figure (e.g., \textit{``Do the upper part of the tree''}). Others relied exclusively on spatial information (e.g., \textit{``Finish the left figure, I'll do the right one''}) or global semantics (e.g., \textit{``Work on the man, I work on the tree''}). For example, participant 11 used voice commands to partition the workspace with the robot, a strategy also reported to improve robot predictability (Table~\ref{tab:preferred_condition}). Less common requests referred to the semantics of specific body parts (e.g., \textit{``Put the foot at the person''}), geometric shapes (e.g., \textit{``Put the triangles of the tree''}), or colors (e.g., \textit{``Put the objects of warm color''}), all of which were correctly interpreted.
Overall, except for the two participants discussed above, all users preferred condition A (Table~\ref{tab:preferred_condition}), citing a greater sense of control, improved robot predictability, more active participation, and a more enjoyable collaboration.

\vspace{-0.2cm}

\section{Discussion and Conclusion}
In this paper, we presented a role allocation framework that leaves the human free to contribute to HRC tasks, combining LLM-based interpretation of user instructions with entropy-based online monitoring of human activities to support intuitive and flexible collaboration.
The experimental results indicate increased \textit{human agency}: participants reported feeling more involved, more in control of the collaboration, and less constrained than under the baseline condition. Allowing users to express verbal preferences and directly observe their impact on the robot's decisions also improved the \textit{intuitiveness} of the interaction. In particular, participants found the robot's behavior easier to predict, whereas in the baseline condition nearly 70\% either failed to understand the allocation strategy or did not pay attention to it, passively following the assigned tasks.
The preference-based condition also resulted in significantly lower frustration levels, suggesting that giving users greater initiative does not increase the interaction burden but instead makes the collaboration more natural and satisfying. The wide variability in the execution strategies adopted by participants, together with the absence of significant differences in perceived workload, further supports the idea that, when no clear distinction exists between activities better suited to the human or the robot, users should be free to decide how to contribute rather than being constrained by a rigid allocation strategy.
Moreover, the proposed framework significantly reduced both task execution time and human idle time, indicating a more fluent collaboration with less passive waiting.
Finally, the framework automatically detects execution anomalies and triggers corrective robot interventions when needed. This closes the loop between activity monitoring, role allocation, and task execution, enabling the robot to recover from human errors while maintaining task progress.
%
%
The current LLM-based instruction handling is tailored to the Tangram assembly task, where each piece can be uniquely identified through simple semantic attributes. Extending the approach to real-world assembly scenarios involving less distinguishable components will make both instruction interpretation and the mapping of user requests to task model activities more challenging.
Moreover, the adopted task model assumes that each activity is executed at most once and that dependencies are purely enabling. Supporting richer task representations, including more complex dependency structures, alternative execution paths, and reversible decisions, will further challenge both the LLM reasoning process and the design of effective prompting strategies.
The entropy-based monitoring also depends on the reliability of the perception pipeline and the availability of representative activity references. In more complex assembly scenarios, multiple execution strategies may correctly accomplish the same task while deviating from the reference interaction patterns, potentially reducing recognition accuracy.
Another limitation is that user requests or actions performed while the robot is executing an activity cannot influence its behavior immediately and are only considered at the next allocation step. Similarly, the current motion planner minimizes execution time without explicitly accounting for human motion. Incorporating human-aware planning, such as hand avoidance or trajectory generation informed by the ongoing human activity, could further improve collaboration fluency.
Finally, integrating predictive models of user behavior could enable the robot to anticipate likely human actions and optimize role allocation over a longer planning horizon while remaining responsive to unexpected behaviors and changing user preferences.
%

\vspace{-0.2cm}

\ifCLASSOPTIONcaptionsoff
  \newpage
\fi

\bibliographystyle{IEEEtran}
\bibliography{biblio}

\end{document}